\documentclass{article}

    \PassOptionsToPackage{numbers, compress}{natbib}

    \usepackage[preprint]{neurips_data_2024}

\usepackage[utf8]{inputenc} %
\usepackage[T1]{fontenc}    %
\usepackage{hyperref}       %
\usepackage{url}            %
\usepackage{booktabs}       %
\usepackage{amsfonts}       %
\usepackage{nicefrac}       %
\usepackage{microtype}      %
\usepackage{xcolor}         %
\usepackage{xspace}
\usepackage{colortbl}
\usepackage{amsmath}
\definecolor{mygray}{gray}{.9}
\definecolor{mygray2}{gray}{.8}
\definecolor{myblue}{rgb}{0.8, 0.8, 1.0}
\usepackage{multirow}
\usepackage{caption}
\usepackage{bbding}
\usepackage{tikz}
\usepackage{subcaption}
\usepackage{makecell}
\usepackage{mdframed}
\makeatletter\renewcommand\paragraph{\@startsection{paragraph}{4}{\z@}
  {.15em \@plus1ex \@minus.2ex}{-.5em}{\normalfont\normalsize\bfseries}}\makeatother
\newcommand{\app}{\raise.17ex\hbox{$\scriptstyle\sim$}}
\makeatletter
\DeclareRobustCommand\onedot{\futurelet\@let@token\@onedot}
\def\@onedot{\ifx\@let@token.\else.\null\fi\xspace}

\def\eg{\emph{e.g}\onedot} 

\def\ie{\emph{i.e}\onedot}

\def\etc{\emph{etc}\onedot} 
\def\vs{\emph{vs}\onedot}

\def\etal{\emph{et al}\onedot}
\makeatother

\title{What If We Recaption \textsf{Billions} of Web Images \\ with \textit{\textsf{LLaMA-3}}?}

\author{%
  Xianhang Li$^{\star 1}$  \,  
  Haoqin Tu$^{\star 1}$ \,
  Mude Hui$^{\star 1}$ \, Zeyu Wang$^{\star 1}$ \,  Bingchen Zhao$^{\star 2}$ \, Junfei Xiao$^3$ \vspace{0.1em} \\   \textbf{Sucheng Ren$^3$} \, \textbf{Jieru Mei$^3$} \, \textbf{Qing Liu$^4$} \, \textbf{Huangjie Zheng$^5$} \, \textbf{Yuyin Zhou$^1$} \, \textbf{Cihang Xie$^1$}
  \vspace{.3em}
  \\\small $^{\star}$equal technical contribution\vspace{.5em} \\
  $^1$UC Santa Cruz \qquad $^2$University of Edinburgh \qquad  $^3$JHU  \qquad  $^4$Adobe \qquad  $^5$UT Austin
}

\begin{document}

\maketitle

\begin{figure}[h]
    \centering
    \vspace{-2.8em}
    \includegraphics[width=0.88\linewidth]{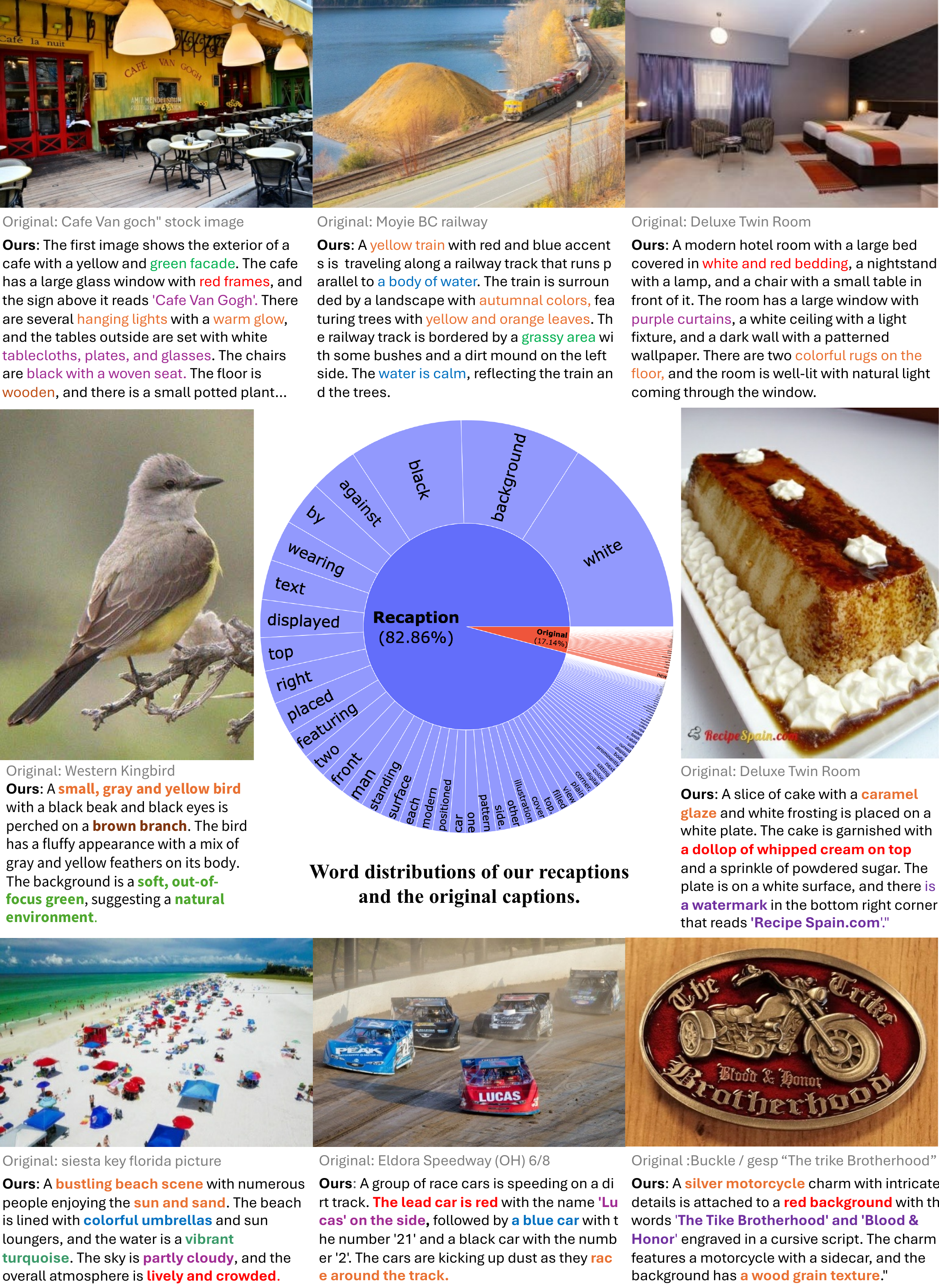}
    \vspace{-.5em}
    \caption{Examples of the original caption and our recaption in DataComp-1B, and word distributions.}
    \label{fig:teaser}
    \vspace{-1.5em}
\end{figure}

\newpage
\begin{abstract}
Web-crawled image-text pairs are inherently noisy. Prior studies demonstrate that semantically aligning and enriching textual descriptions of these pairs can significantly enhance model training across various vision-language tasks, particularly text-to-image generation. However, large-scale investigations in this area remain predominantly closed-source. 
Our paper aims to bridge this community effort, leveraging the powerful and \textit{open-sourced} LLaMA-3, a GPT-4 level LLM.
Our recaptioning pipeline is simple: first, we fine-tune a LLaMA-3-8B powered LLaVA-1.5 and then employ it to recaption \app1.3 billion images from the DataComp-1B dataset. Our empirical results confirm that this enhanced dataset, Recap-DataComp-1B, offers substantial benefits in training advanced vision-language models.  For discriminative models like CLIP, we observe enhanced zero-shot performance in cross-modal retrieval tasks.  For generative models like text-to-image Diffusion Transformers, the generated images exhibit a significant improvement in alignment with users' text instructions, especially in following complex queries.
Our project page is \url{https://www.haqtu.me/Recap-Datacomp-1B/}.
\vspace{-.4em}
\end{abstract}

\section{Introduction}
The exponential growth in data availability is one of the most paramount factors in driving the monumental successes of deep learning over the past decade \cite{deng2009imagenet,lin2014microsoft,cc3m,laion,gadre2023datacomp,dfn}. Typically, this data is sourced through web crawling with simple filtering mechanisms in place. While such an approach has facilitated large-scale data collection, exemplified by collections like LAION-400M~\cite{laion} and LAION-5B~\cite{laion} with billions of image-text records, it has inadvertently compromised data quality. As illustrated in Figure \ref{fig:teaser}, these internet-crawled image-text pairs frequently exhibit misalignments between images and their corresponding textual content, and often, the textual descriptions are brief and lack detailed information.

To mitigate the noise present in web-crawled data, enhancements through post-processing---implemented via human-in-the-loop systems \cite{sun2023aligning, yu2023rlhf} or automated pipelines \cite{laion,blip,blip2}---are crucial, which help to train the advanced vision-language foundation models. Notably, both the \emph{close-sourced} DALL-E 3~\cite{dalle3} and SORA~\cite{sora} incorporate advanced captioning techniques to re-label their training datasets, a crucial step highlighted in their technical reports.  Despite various efforts to open-source and replicate these methodologies \cite{sharegpt4v,blip,blip2,llava,yu2023capsfusion,Laclip,rotstein2023fusecap}, the community continues to face significant challenges in accessing high-quality, well-aligned image-text data at scale (\eg, at the billion level) for training advanced vision-language foundation models.

This paper endeavors to contribute to this community initiative, inspired specifically by the release of LLaMA-3~\cite{llama3}, a model demonstrating GPT-4-level capabilities across a variety of linguistic tasks. Additionally, recent studies have shown that leveraging LLaMA-3 can significantly enhance model performance on vision-language tasks \cite{llavanext,minicpm}, comparable to those achieved by GPT-4V~\cite{gpt4}. In response, we employ LLaMA-3 to develop our advanced captioner model. Our approach is straightforward: we first train a LLaMA-3-powered Llava model to act as an image captioner, which is then utilized to recaption the entire DataComp-1B dataset. As depicted in Figure \ref{fig:teaser}, the resulting dataset, dubbed Recap-DataComp-1B, features enhanced textual descriptions and improved alignment with corresponding images, clearly surpassing its web-crawled counterparts. These quality enhancements are further quantitatively verified in Section \ref{sec:data_quality}.

Comprehensive evaluations highlight the significant improvements that Recap-DataComp-1B contributes to the training of advanced vision-language foundation models. Notably, this dataset enables CLIP models to achieve significant enhancements in their zero-shot cross-modal retrieval capabilities. It also enhances the alignment between generated images and text instructions in text-to-image generative models pre-trained on our dataset. We hope that the release of Recap-DataComp-1B will catalyze further developments in advanced vision-language foundation models, particularly encouraging the development within the open-source community.

\section{Related works}
\paragraph{Vision-Language Foundation Models.} CLIP~\cite{clip} is one of the pioneering foundation models to connect image and text. By training on millions, and even billions, of image-text pairs \cite{cc3m,desai2021redcaps,dfn,gadre2023datacomp,schuhmann2022laion,laion,sharma2018conceptual,srinivasan2021wit}, CLIP markedly showcases excessively strong zero-shot capacities, and furthermore, lays the cornerstone for building more advanced vision-language foundation models \cite{flamingo,blip,blip2,simvlm,llava,llavanext,internvl,Qwen-VL,minicpm}. Apart from discriminative vision-language understanding, 
text-to-image generation models \cite{ding2021cogview,nichol2021glide,dalle3,peebles2023scalable,dalle2,ramesh2021zero,ldm,imagen,parti} have transformed the field of AI-generated content, facilitating the creation of high-quality images from natural language descriptions.

\paragraph{Enhancing Image-Text Data.} Web-crawled image-text 
data~\cite{laion,gadre2023datacomp,dfn} commonly face the problems of image-text misalignment and the low-quality of textual descriptions. 
Typically, there are two popular ways for improving the quality of these image-text pairs: 1) \emph{data filtering} removes misaligned image-text pairs using various methods such as cleaning strategies~\cite{schuhmann2022laion, gadre2023datacomp, metaclip}, pretrained models~\cite{blip,laion, gadre2023datacomp}, and human-assisted systems~\cite{sun2023aligning, yu2023rlhf,zhang2023compress}; 2) \emph{data recaptioning} improves the textual quality of image-text pair via generating new captions, which is the focus of this paper.
To recaption data, LaCLIP~\cite{Laclip} utilizes large language models (LLMs) like ChatGPT to rewrite the original captions; Nguyen \etal \cite{nguyen2024improving} employ BLIP2 \cite{blip2} to recaption images. More recently, advanced large multimodal models have been applied to further enhance the quality of image captioning.  For example, ShareGPT4V~\cite{sharegpt4v} employs GPT-4V to create highly descriptive captions from carefully crafted prompts and corresponding image inputs; the resulting dataset has significantly benefited the training of various models \cite{chen2024pixart,zhang2024longclip,chu2024mobilevlm,lin2024moellava,fei2024dimba}. However, scaling such prompting with GPT-4V to billions of records is less practical, as it will drastically increase the monetary cost (of intensively calling OpenAI APIs) by more than 10,000$\times$.

Our paper mostly follows the approach presented in \cite{chen2024pixartalpha,lu2023pangudraw,zhang2024longclip,chu2024mobilevlm}, where advanced open-source multimodal models like LLaVA \cite{llava} are employed for recaptioning purposes. However, our approach is distinguished by two major aspects: 1) we strongly enhance the LLM module in LLaVA, \ie, building with LLaMA-3; and 2) our recaptioning efforts are executed on a billion-scale dataset.

\section{Recaptioning Pipeline}
\label{sec:method}

Our recaptioning pipeline is centered around the advanced LLM
LLaMA-3 \citep{llama3},
which achieves exceptionally strong performance in language understanding, reasoning, code generation, math problems, \etc~\cite{chiang2024chatbot,mediumllama3perf}. Specifically, we utilize the LLaVA framework \cite{llava} to fully harness its capabilities for 
visual understanding. We describe the detailed training procedures below.

\subsection{Model details}

\begin{figure}[t]
    \centering
    \vspace{-1em}
    \includegraphics[width=.99\linewidth]{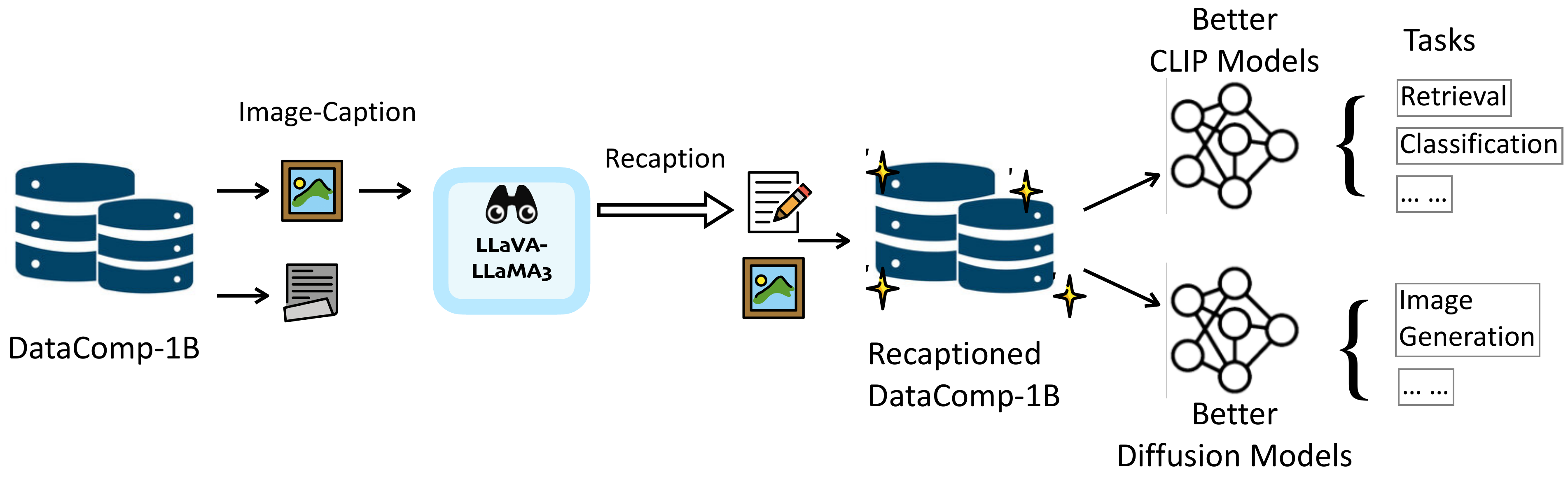}
    \vspace{-.5em}
    \caption{The illustration of our recaptioning pipeline on DataComp-1B. We use LLaMA-3-powered LLaVA to reception images, which enables us to train stronger CLIP models and Text-to-Image Diffusion models.}
    \label{fig:recap_pipeline}
    \vspace{-.9em}
\end{figure}

\paragraph{Model Configuration.} 
We follow the setup of LLaVA-1.5~\cite{liu2023improved} to build our captioner model, except that we use LLaMA-3-8B as the language decoder because of its superior performance. 
The visual branch of CLIP ViT-L/14~\cite{radford2021learning} is used as the vision encoder. Two trainable MLP layers are employed on top of the vision encoder to project visual features into the language embedding space.

\paragraph{2-Stage Training.}

We also follow LLaVA-1.5~\cite{liu2023improved} for model training. Essentially we conduct instruction-tuning on the pre-trained LLM with its original
auto-regressive training objective. In the first stage, only the projection MLP is trained;
in the second stage, we fine-tune both the projection MLP and the language decoder. Note that the vision encoder remains frozen all the time. Following the protocols in LLaVA~\cite{liu2023improved}, 558k image-text pairings filtered from LAION~\cite{schuhmann2022laion}, CC~\cite{cc3m},and SBU~\cite{ordonez2011im2text} are used as training data in the first stage; then 665k instructions-following data from
LLaVA-1.5~\cite{liu2023improved}, containing image-grounded conversation, image descriptions, and image-based complex reasoning tasks, are used for the second stage of training. \emph{To help our model generate higher-quality captions, we use the image-text pairs from HQ-Edit dataset~\cite{hqedit2024} for further tuning}.

\paragraph{Evaluations.} To probe the visual understanding and reasoning ability of our LLaVA-1.5-LLaMA3-8B model, we opt for two comprehensive multi-modal evaluation benchmarks, MMMU~\cite{yue2023mmmu} and MM-Vet~\cite{yu2024mm}. These benchmarks assess a broad range of capabilities such as recognition, spatial awareness, OCR, knowledge, and language generation.
As reported in Table \ref{tab:test_llava_llama3}, on both benchmarks, our LLaVA-1.5-LLaMA3-8B model surpasses the vanilla LLaVA-1.5-7B model by a significant margin. These results also substantially beat the considerably larger LLaVA-1.5-13B model, clearly demonstrating the superior visual understanding and reasoning ability of our model.

\begin{table}[t!]
    \centering
    \caption{Performance comparison of LLaVA.}
    \label{tab:llava_comp}
    \begin{tabular}{ccccc}
    \toprule
    Model & LLaVA-1.5-7B & LLaVA-1.5-13B & LLaVA-1.5-LLaMA3-8B (\textbf{ours}) &\textcolor{gray}{GPT-4V}  \\
    \midrule
    MMMU & 33.6 & 36.4 & \textbf{45.2} & \textcolor{gray}{56.8}  \\
    MM-Vet & 33.9 & 36.3 & \textbf{37.8} & \textcolor{gray}{44.6}  \\
    \bottomrule
    \end{tabular}
    \vspace{-.5em}
    \label{tab:test_llava_llama3}
\end{table}

\subsection{Recaptioning DataComp-1B}
With this advanced LLaVA model, we next use it to generate captions in a scalable and detailed manner, given the visual input, and the following text prompt: 
$$\textit{"Please generate a detailed caption of this image. Please be as descriptive as possible."} $$
As for the dataset, we opt for DataComp-1B ~\cite{gadre2023datacomp}, a widely accessible, large-scale vision-language dataset comprising \app1.3 billion web-crawled image-text pairs.
To ensure its quality, DataComp-1B is already a curated subset from a much larger collection of 12.8 billion image-text pairs and has been subjected to rigorous preprocessing which includes safety checks, deduplication, CLIP score filtering, and image-based filtering.  Despite these efforts, the quality of the original captions in DataComp-1B still exhibits relatively low quality. 

We apply our well-trained LLaVA-1.5-LLaMA3-8B model to recaption the entire DataComp-1B dataset. Specifically,
captions are generated auto-regressively via greedy decoding, with the maximum output token length set to 128.
We term this newly recaptioned dataset \textit{Recap-DataComp-1B}.

\section{Analyzing Recap-DataComp-1B}
\label{sec:data_quality}
This section collects and presents a quantitative analysis of our generated captions on DataComp-1B. We primarily focus on two aspects: 1) the inherent features of the captions, including word distributions and average lengths; and 2) the semantic quality of the captions, evaluated in terms of the matching similarity between images and captions and the inherent textual quality of the captions.

\subsection{Word \& Length Distribution}
\label{sec:data_length}
\begin{figure}[t!]
    \centering
    \vspace{-1em}
    \includegraphics[width=\linewidth]{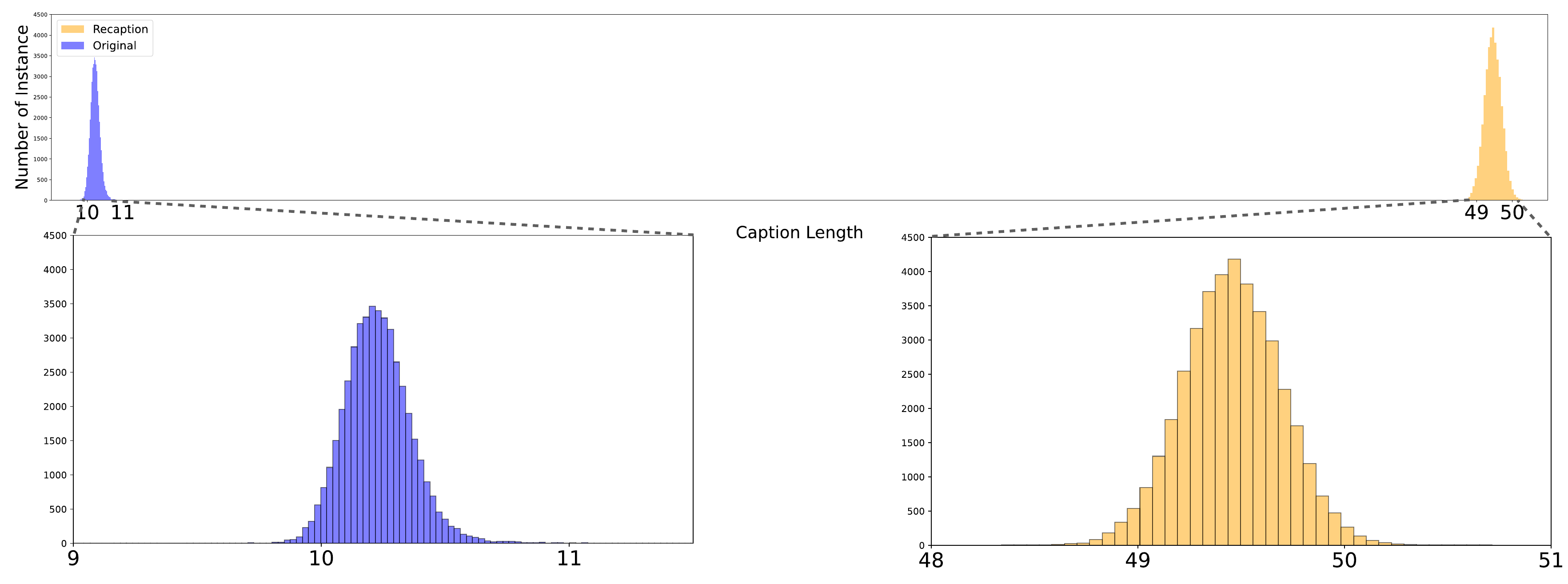}
    \vspace{-1.5em}
    \caption{Average length distributions of both the original captions and our recaptioned data in DataComp-1B.}
    \label{fig:len_dis}
    \vspace{-1em}
\end{figure}

We begin our analysis by comparing the word frequency distributions between our recaptioned content and that from the original DataComp-1B, as illustrated in Figure~\ref{fig:teaser}, analyzing a randomly sampled subset of approximately 0.35 billion instances. 
Our findings reveal that the recaptioned content displays a considerably richer vocabulary, capturing 82.86\% tokens of the word collections from both ours and the original caption data. Additionally, there is a noticeable variety in the usage of nouns and adjectives in our captions (\eg, ``white'' and ``background''). 
We argue that this increased lexical diversity is a direct consequence of the extended length of our data. 
We thus present the distribution of caption lengths in Figure~\ref{fig:len_dis} to highlight this difference. On average, our recaptioned data demonstrates a longer sequence length of 49.43, whereas the original DataComp captions have a much shorter length of 10.22. 
These observations validate that our Recap-DataComp-1B surpasses the original DataComp-1B version in terms of both caption length and diversity.

\subsection{GPT-4V \& CLIP Evaluations}
\label{sec:GPT-4V Evaluations}
Next, we evaluate the semantic quality of recaptioned content using two models: 1) CLIP \cite{clip}, which measures the semantic similarity between captions and images, and 2) GPT-4V~\cite{gpt4v}, which assesses the fluency and alignment of captions with the given images.

For the CLIP evaluation, we analyze a subset of 180,000 image-text pairs. Interestingly, we note that, when using the standard CLIP-large model with \app428M parameters for this measurement,
our recaptioned content performs just comparably to the original captions (49.57 \vs 50.43). We attribute this result primarily to the limitations of the standard CLIP model, which is trained on `short' captions and may inadequately capture the nuances in semantic similarity for longer captions.
To probe deeper into semantic alignment between long captions and images, we utilize the LongCLIP-Large model \cite{zhang2024longclip}, which is specifically fine-tuned to handle longer captions. With this setup, the LongCLIP score of our newly generated caption impressively attains 89.91, nearly $9\times$ higher than the LongCLIP score of the original DataComp captions (\ie, only 10.09).

In addition, to evaluate both the textual quality and the alignment of the captions with their corresponding images, we randomly select 10,000 instances for GPT-4V~\cite{gpt4v} evaluation, employing the prompting strategy outlined below (\textcolor{red}{CAPTION} is the textual input), as per~\cite{padlewski2024vibe,vhelm_initial}. 

\begin{mdframed}[backgroundcolor=gray!15]
\textbf{GPT-4V Evaluation Instruction:}\\{[}Image Caption]\\\textcolor{red}{CAPTION} \\ \\Rate whether the caption is of high-quality and fluent and correctly matches the given image. The rating should be 1-5, where 1 is incorrect and not fluent at all, and 5 is correct and very fluent. Try to just give a numerical rating.\\ \\Your response should be in the format:\\Rating: (int)
\end{mdframed}

We can observe that our recaptioned content achieves markedly superior ratings, registering an average rating increase of 0.43 (from 3.71 to 4.14). Together with the findings from Section \ref{sec:data_length}, this confirms the superior quality of our newly generated captions, in terms of length, vocabulary diversity, semantics, and image-text alignment.

\section{CLIP}

CLIP~\cite{clip} stands as a widely utilized vision-language model, where an image encoder and a text encoder are jointly trained to predict correct matches across entire batches of image-text pairs.
In this section, we delve into the advantages of training CLIP models with our Recap-DataComp-1B dataset.
We anticipate that CLIP models trained on this dataset will exhibit superior zero-shot cross-modal retrieval capabilities and enhanced text understanding, especially with long and complex textual inputs, given the improved quality of our recaptions.

\subsection{Experiment settings}

\paragraph{Training.} 
For reference, we term the CLIP model trained on our Recap-DataComp-1B dataset as Recap-CLIP.
Our training pipeline primarily follows CLIPA~\cite{li2023clipa,li2023clipav2}, which incorporates a two-state training, \ie, a pre-training process with a small image size followed by a fine-tuning stage incorporating a larger image resolution.
We set the text token length to 128 to accommodate the learning of long captions presented in Recap-DataComp-1B.
We conduct experiments using three model scales: S/16, B/16, and L/16, with details listed in Table~\ref{tab:arch}. The AdamW~\cite{adamw} optimizer is used for training.
In the pre-training phase, the model is trained with 2.56 billion samples with a reduced image size of 112, including a warm-up phase involving 51.2 million samples. The batch size and base learning rate are set to 32,768 and 8e-6, respectively.
For the subsequent fine-tuning phase, we increase the image size to 224 and train the model on 128 million samples with a 25.6 million sample warm-up. Here, we adjust the batch size to 16,384 and the learning rate to 4e-7.

\paragraph{Evaluation.} 
The efficacy of Recap-CLIP is gauged via several metrics. We evaluate zero-shot image classification on the ImageNet-1K dataset~\cite{ILSVRC15} and assess zero-shot cross-modal retrieval performance using the validation set of MSCOCO 2014~\cite{lin2014microsoft} and the test set of Flickr30K~\cite{flickr30k}\footnote{We employ the widely used Karpathy split~\cite{karpathy2015deep} of MSCOCO and Flickr30K.}, following the established practices~\cite{clip,li2023clipa,zhai2023sigmoid,zhai2022lit}.

We present our results from three aspects. First, we explore the impacts of differing mix ratios between original captions and our enhanced recaptions on CLIP performance. Next, we analyze the effects of enlarging the size of the CLIP text encoder.
Lastly, we investigate the text understanding capability of our Recap-CLIP, via testing on VG-Attribute~\cite{yuksekgonul2022aro}, which evaluates attributes understanding ability, and Urban1K~\cite{zhang2024longclip}, which tests the model's ability to handle long text.

\begin{table}[t!]
 	\caption{\textbf{Recap-CLIP model configurations} used in our paper.
	}
    \centering
    \resizebox{.8\linewidth}{!}{
	\begin{tabular}{l|c|ccc|ccc|ccc}
          \toprule
		& Embed &    \multicolumn{3}{c|}{Vision Transformer} & \multicolumn{3}{c|}{Text Transformer} & \multicolumn{3}{c}{\# params (M)} \\
		model & dim  & layers & width & heads  & layers & width & heads & vision & text & total \\ 
		\midrule
            S/16  & 384 & 12 & 384 &6 & 12 & 384 &6 & 22 & 33 & 55\\
		B/16  & 512  & 12 & 768 & 12 & 12 & 512 & 8 & 86 & 53 & 141\\
		L/16  & 768  & 24 & 1024 & 16 & 12 & 768 & 12 & 303 & 109 & 414\\
		H/14  & 1024 & 32 & 1280 & 16 & 24 & 1024 & 16 & 631 & 334 & 967\\
  \bottomrule
	\end{tabular}
    }
	\label{tab:arch}
 \vspace{-.1em}
\end{table}

\begin{table*}[t!]
    \centering
    \caption{\textbf{Train with mixed captions. We choose Recap-CLIP-B/16 for this ablation.} Larger $p$ represents a higher ratio of the original caption. We report top-1 zero-shot classification accuracy on ImageNet-1K and top-1 recall for retrieval tasks. *concat: Concat two types of captions.}
    \label{tab:ablation_mixed_captions}
    \resizebox{.7\linewidth}{!}{
    \begin{tabular}{c|c|c|c|c|c}
    \toprule
      \multirow{2}{*}{\textbf{mixed ratio $p$}} &\textbf{ImageNet-1K} &\multicolumn{2}{c}{\textbf{COCO  R$@$1}} & \multicolumn{2}{c}{\textbf{Flickr30K  R$@$1}} \\
      \cmidrule(r){2-6}   
     &Validation 
     &\multicolumn{1}{c}{I$\rightarrow$T}
     &\multicolumn{1}{c}{T$\rightarrow$I}
     &\multicolumn{1}{c}{I$\rightarrow$T}
     &\multicolumn{1}{c}{T$\rightarrow$I} \\
     \midrule
    0.0 & 36.0 & 53.0 & 34.1 &74.1  & 53.5  \\
    0.1 & 58.4 & 60.9 & 40.5 &83.9 &65.5  \\
    0.2 & 62.5 & 61.7 & 41.4 &85.8  &65.7 \\
    0.3 & 65.1 & \textbf{62.7} & 42.6 &86.3 & 67.0 \\
    0.4 & 66.7 & 62.6 & 42.5 &\textbf{87.4}  &\textbf{67.7} \\
    0.5 & 67.2 & 61.9 & \textbf{42.7} &85.9 &66.7  \\
    0.6 & 68.0 & 62.2 & 42.4 &86.0 &67.4\\
    0.7 & 68.4 & 60.7 & 42.3 &86.3 &66.9  \\
     \rowcolor{mygray} 0.8 & 69.2 & 61.5 & 42.2 &85.2 &66.9  \\
    0.9 & 69.2 & 60.6 & 41.1 &86.0   &65.7\\
    1.0 & \textbf{69.7} & 57.3 & 37.7 &84.2  &63.0 \\
    \midrule
    *concat & 43.3 & 57.8 & 35.6 &80.2  &56.4 \\
    \bottomrule
    \end{tabular}}
    \vspace{-1em}
\end{table*}

\subsection{Training with Mixed Captions}
As pointed out by DALL-E 3~\cite{dalle3}, blending both the briefgenuine captions and the long informative generated captions can effectively prevent the model from unwanted overfitting to recaption data. Therefore, we hereby first study the effect of varying mix ratios between the original captions and our recaptions on the training of the Recap-CLIP B/16 model, as detailed in Table \ref{tab:arch}. Specifically, for each sample in a training batch, we randomly sample the original caption with probability \( 0 \leq p \leq 1 \) and our captions with probability \( 1 - p \), referring to the mixed ratio:
\[
\text{Caption} = 
\begin{cases} 
\text{Original} & \text{with probability } p \\
\text{Recaption} & \text{with probability } 1 - p 
\end{cases} 
\]
This strategy ensures that each batch contains a mixture of our recaption and the original captions controlled by probability $p$. 
The randomness allows each sample to encounter different captions across training epochs, potentially enhancing the model's generalization.

\paragraph{Main results.}
Our findings are summarized in Table~\ref{tab:ablation_mixed_captions}.
We observe that reducing the mixed ratio $p$ (\ie, increasing the proportion of our recaption data) initially leads to an improvement followed by a decline in cross-modal retrieval performance.
This initial improvement suggests that high-quality recaptioned data effectively enhances contrastive learning. However, the subsequent decrease indicates that the original captions from DataComp-1B provide necessary training regularization, preventing the model from overly adapting to the specific qualities of the recaption data.
Interestingly, we also observe that the performance of CLIP is relatively insensitive to certain variations in the mix ratio $p$, as evidenced by the consistent enhancement over the baseline (\ie $p$=1.0) across all four different cross-modal retrieval metrics when varying $p$ from 0.2 (80\% recaption data) to 0.9 (10\% recaption data). 
For instance, setting $p$ at 0.9 and 0.2 both yields a similar performance enhancement of $\sim$3.5\%, with the peak performance occurring at $p$=0.5, which delivers a substantial \app5\% boost. 

But meanwhile, we note that incorporating our recaptions (negatively) affects the zero-shot classification task, exemplified by the consistent performance degradation across varying $p$ values from 0 to 0.9. The phenomenon is similarly observed in the 
recent work~\cite{zhang2024longclip} where they note directly fine-tuning on long text can significantly hurt the CLIP performance and therefore propose several techniques for enhancing learning with long texts. 
In this study, given our primary focus is on assessing the quality of Recap-DataComp-1B, we choose the ratio $p=0.8$ to strike a promising balance between the classification performance (\ie, only marginally drops 0.5\%) and the cross-modal retrieval performance (\ie, with a significant 3.4\% boost on average) for later ablations.

\begin{table*}[t!]
    \centering
    \caption{\textbf{Train with larger text encoder.} We set $p=0.8$ for recaption-based models. We report zero-shot top-1 accuracy on ImageNet-1K and top-1 recall on COCO and Flickr30K.
    }
    \label{tab:ablation_larger_text_encoder}
    \resizebox{\linewidth}{!}{
    \begin{tabular}{c|c|c|l|l|l|l|l}
    \toprule
       \multirow{2}{*}{\textbf{vision encoder}}&\multirow{2}{*}{\textbf{text encoder}} &\multirow{2}{*}{\textbf{re-caption}} &{\textbf{ImageNet-1K}} & \multicolumn{2}{c}{\textbf{COCO R$@$1}} &\multicolumn{2}{c}{\textbf{Flickr30K R$@$1}}\\
     \cmidrule(r){4-8}   
     &&&Validation
     &\multicolumn{1}{c}{I$\rightarrow$T} 
     &\multicolumn{1}{c}{T$\rightarrow$I} 
     &\multicolumn{1}{c}{I$\rightarrow$T} 
     &\multicolumn{1}{c}{T$\rightarrow$I} \\
     \midrule
     \multirow{4}{*}{S/16} &small &\XSolidBrush &60.7  &49.2 &30.1 &73.5 &53.3 \\
      &small &\cellcolor{mygray}\CheckmarkBold  &\cellcolor{mygray}60.2  
      &\cellcolor{mygray}53.7 %
      &\cellcolor{mygray}34.3 %
      &\cellcolor{mygray}78.6 %
      &\cellcolor{mygray}57.9 \\ %
      &\emph{base} &\cellcolor{mygray}\CheckmarkBold &\cellcolor{mygray}61.7 \textcolor{red}{$_{+1.5\%}$}
      &\cellcolor{mygray}56.4 \textcolor{red}{$_{+2.7\%}$}
      &\cellcolor{mygray}34.8 \textcolor{red}{$_{+0.5\%}$}
      &\cellcolor{mygray}79.7 \textcolor{red}{$_{+1.1\%}$} 
      &\cellcolor{mygray}59.1 \textcolor{red}{$_{+1.2\%}$} \\
     \midrule
     \multirow{4}{*}{B/16} &base &\XSolidBrush &69.7 &57.3 &37.7 &84.2 &63,0\\
      &base &\cellcolor{mygray}\CheckmarkBold &\cellcolor{mygray}69.2 
      &\cellcolor{mygray}61.5 %
      &\cellcolor{mygray}42.2 %
      &\cellcolor{mygray}85.2 %
      &\cellcolor{mygray}66.9 \\ %
      &\emph{large} &\cellcolor{mygray}\CheckmarkBold &\cellcolor{mygray}69.8 \textcolor{red}{$_{+0.6\%}$}
      &\cellcolor{mygray}62.9 \textcolor{red}{$_{+1.4\%}$} 
      &\cellcolor{mygray}42.8 \textcolor{red}{$_{+0.6\%}$} 
      &\cellcolor{mygray}86.7 \textcolor{red}{$_{+1.5\%}$} 
      &\cellcolor{mygray}67.3 \textcolor{red}{$_{+0.4\%}$} \\  
      \midrule
      \multirow{4}{*}{L/16} &large &\XSolidBrush &74.1 &60.2 &41.9 &86.0 &68.5 \\
      &large &\cellcolor{mygray}\CheckmarkBold &\cellcolor{mygray}73.8 
      &\cellcolor{mygray}64.3 %
      &\cellcolor{mygray}46.1 %
      &\cellcolor{mygray}88.3 %
      &\cellcolor{mygray}70.5 %

 \\
      &\emph{huge} &\cellcolor{mygray}\CheckmarkBold 
      &\cellcolor{mygray}74.2 \textcolor{red}{$_{+0.4\%}$}
      &\cellcolor{mygray}66.0 \textcolor{red}{$_{+1.7\%}$} 
      &\cellcolor{mygray}46.6 \textcolor{red}{$_{+0.5\%}$} 
      &\cellcolor{mygray}89.9 \textcolor{red}{$_{+1.6\%}$} 
      &\cellcolor{mygray}72.7 \textcolor{red}{$_{+2.2\%}$} 
 \\ 
          \bottomrule
    \end{tabular}}
    \vspace{-.5em}
\end{table*}

\subsection{Training with Larger Text Encoder}
We hereby investigate how the size of the text encoder affects models trained on a mixture of the original captions and our recaptions (with $p=0.8$).
Specifically, we keep the architectural configuration of the vision branch as in Table~\ref{tab:arch} and only twitch the text encoder. 
For instance, in the case of the S/16 model, we change from a smaller text encoder with 33M parameters to a larger, base-sized one with 53M parameters.

\paragraph{Main Results} Our results, as shown in Table \ref{tab:ablation_larger_text_encoder}, highlight that 
enlarging the text encoder can further enhance performance across all model scales.
The average improvement for adopting a larger text encoder in retrieval tasks is 1.4\%, 1.0\%, and 1.5\% for small, base, and large models, respectively, suggesting that larger text encoders can help the CLIP model learn better from semantically rich captions.

Moreover, we re-assess the balanced ratio of recaption data using a larger text encoder. Specifically, we gradually increase the ratio of recaption data from 20\% to 50\%, utilizing the Recap-CLIP-B/16 model with the \textit{large text encoder}. The results are presented in Table~\ref{tab:large_text_encoder_mixed_ratio}. Compared to the prior results where an optimal ratio is achieved at $p=0.8$, using a larger text encoder can further push this optimal ratio to $p=0.6$. In other words, this result concludes that, compared to the vanilla version, a stronger cross-modal retrieval performance can be achieved if 1) more recaptions are used and 2) a larger text encoder is used.

\subsection{More evaluations on text understanding}
Recent works demonstrate that CLIP models suffer from poor long context understanding and delicate attribute understanding~\cite{yuksekgonul2022aro,zhang2024longclip}. Given the long, enriched, and better-aligned captions, we expect Recap-CLIP to exhibit better text understanding capability. Thus, we evaluate our Recap-CLIP model on two benchmarks: (1) Urban1K~\cite{zhang2024longclip}, a long-caption image-text retrieval benchmark that contains 1k urban images and corresponding GPT-4V captions; (2) VG-Attribution~\cite{yuksekgonul2022aro}, a modified version of Visual Genome~\cite{krishna2017visual} to test model abilities to attribute properties to objects. The results are shown in Tab.~\ref{tab:longclip_evaluation_for_clip}. 

We observe consistent significant improvement if the model is trained on our Recap-Datacomp-1B dataset. For both text-to-image and image-to-text retrieval on Urban-1K dataset, our Recap-CLIP models surpass the vanilla baseline by at least 19\% and sometimes up to an astonishingly high 36\%. On the VG-attribution dataset, it is worth noting that our Recap-CLIP brings a performance boost very close to that of the NegCLIP fine-tuning~\cite{yuksekgonul2022aro} (\eg $\sim$9\% \vs 10\%), a lightweight downstream fine-tuning process designed to boost CLIP ability to understand attribute and order. Nonetheless, it is noteworthy that our Recap-CLIP is naturally equipped with better text understanding ability, without any specific targeted fine-tuning, indicating the importance of better captions in web-scale data.
\begin{table*}[t!]
    \centering
        \centering
        \caption{\textbf{Larger text encoder with different mixed ratios.} We choose Recap-CLIP-B/16 with \textit{large} text encoder for this ablation.}
        \label{tab:large_text_encoder_mixed_ratio}
        \resizebox{0.63\linewidth}{!}{
        \begin{tabular}{c|c|c|c|c|c}
            \toprule
            \multirow{2}{*}{\textbf{mixed ratio $p$}} & \textbf{ImageNet-1K} & \multicolumn{2}{c}{\textbf{COCO  R$@$1}} & \multicolumn{2}{c}{\textbf{Flickr30K  R$@$1}} \\
            \cmidrule(r){2-6}
            & Validation & \multicolumn{1}{c}{I$\rightarrow$T} & \multicolumn{1}{c}{T$\rightarrow$I} & \multicolumn{1}{c}{I$\rightarrow$T} & \multicolumn{1}{c}{T$\rightarrow$I} \\
            \midrule
            0.5 &68.5&64.3&43.4&86.8&67.8\\
            \cellcolor{myblue}0.6 &\cellcolor{myblue}69.2&\cellcolor{myblue}\textbf{64.4}&\cellcolor{myblue}\textbf{43.2}&\cellcolor{myblue}87.5&\cellcolor{myblue}\textbf{68.8}\\
            0.7&69.3&63.2&42.7&\textbf{88.0}&68.2\\
            \cellcolor{mygray}0.8 &\cellcolor{mygray}\textbf{69.8}&\cellcolor{mygray}62.9&\cellcolor{mygray}42.8&\cellcolor{mygray}86.7&\cellcolor{mygray}67.3\\
            \bottomrule
        \end{tabular}}
\end{table*}
\begin{table}[t!]
\centering
\vspace{-0.5em}
\caption{Comparison on the Urban-1K and VG-Attribute benchmark.}
\resizebox{.7\linewidth}{!}{
\begin{tabular}{c|c|lll}
\toprule
  \multirow{2}{*}{\textbf{method}} & \multirow{2}{*}{\textbf{re-caption}} &\multicolumn{2}{c}{\textbf{Urban-1K}} &\textbf{VG}\\
      \cmidrule(r){3-5}
 & &\multicolumn{1}{l}{I$\rightarrow$T} &\multicolumn{1}{l}{T$\rightarrow$I} &Attribute\\
\midrule
OpenAI-CLIP-B/16~\cite{clip} &\XSolidBrush &67.4 &53.3 &62.6 \\ 
OpenCLIP-B/16~\cite{openclip} &\XSolidBrush &62.5 &63.1 &59.9 \\ 
\hline
\multirow{2}{*}{Recap-CLIP-B/16}  &\XSolidBrush &53.2 &50.9 &57.1 \\
 &\cellcolor{mygray}\CheckmarkBold &\cellcolor{mygray}85.0 \textcolor{red}{$_{+31.8\%}$}
&\cellcolor{mygray}87.3 \textcolor{red}{$_{+36.4\%}$} 
&\cellcolor{mygray}66.4 \textcolor{red}{$_{+9.1\%}$}\\
\cmidrule(r){2-5}
\multirow{2}{*}{Recap-CLIP-L/16}  &\XSolidBrush &69.8 &64.6 &60.1 \\
  &\cellcolor{mygray}\CheckmarkBold &\cellcolor{mygray}89.0 \textcolor{red}{$_{+19.2\%}$}
&\cellcolor{mygray}91.8 \textcolor{red}{$_{+27.2\%}$}
&\cellcolor{mygray}66.8 \textcolor{red}{$_{+6.7\%}$}\\
\bottomrule
\end{tabular}
}
\vspace{-.5em}
\label{tab:longclip_evaluation_for_clip}
\end{table}

\section{Text-to-Image Generation}

It has been known to the research community that training with generated (high-quality) pseudo-captions improves text-to-image generative models in terms of generation quality and prompt following ability~\cite{chen2024pixartalpha,chen2024pixart,betker2023improving},
primarily due to the low information and high noise density presented in the original web-crawled captions. Therefore, 
we evaluate the quality of our generated captions by training Text-to-Image (T2I) generative models on Recap-DataComp-1B for further justification.  We expect the enriched information in the generated descriptions to better align the visual content in images, and thus improve the performance of the T2I models.

\textbf{Training.} We adopt Diffusion Transformers (DiT)~\cite{peebles2023scalable} as our T2I model, where the text condition is firstly extracted with a CLIP text encoder~\cite{clip}, and then injected into each DiT block with the cross-attention design. Specifically, we follow the image preprocessing pipeline in DiT~\cite{peebles2023scalable}, where the images are preprocessed to have a square resolution of 256. The model is trained on visual latent extracted using a pretrained auto-encoder with a downsampling ratio of 8~\cite{ldm}. Similar to the setup in previous experiments, the training text consists of a mixture of raw captions from Datacomp-1B, with a specified proportion $p$, and the rest of the captions replaced by refined captions from Recap-Datacomp-1B. Moreover, the training batch size is 2048, and the AdamW optimizer~\cite{adamw} is used with a constant 1e-4 learning rate, without any warm-up schedule or weight decay. We name the resulting model Recap-DiT.

\textbf{Evaluation. } For sampling, we set the classifier-free guidance scale as 10 and use 250 DDPM steps to generate 30k images with captions from MSCOCO and our improved generated captions for zero-shot generation evaluation. We calculate Fréchet Inception Distance (FID)~\cite{heusel2017gans} with the reference images from MSCOCO~\cite{lin2014microsoft} and CLIP score with both OpenAI ViT-B/32 model~\cite{clip} and our own Recap-CLIP ViT-L/16 model, following the established pipeline in prior T2I works~\cite{betker2023improving,yu2022scaling,sauer2023stylegan,kang2023scaling,liu2023insta,zhou2024long,Sauer2023AdversarialDD}. 
Additionally, following the GPT-4V metric introduced in Section~\ref{sec:GPT-4V Evaluations}, we randomly select a subset of 3,000 our generated images for GPT-4V evaluation.

\textbf{Main results. } We report our observations in Tab.~\ref{tab:dit_mix_ratio}. %
Interestingly, when using raw COCO captions to generate 30,000 images for evaluation, the model trained with data integrated with our Recap-Datacomp (for $p<1$) demonstrates a better CLIP score, indicating improved vision-language alignment. However, there is no significant improvement observed in terms of FID. Our hypothesis is that the model adapts to the more informative and descriptive prompts, and could unleash its full potential only when similar informative testing prompts are provided.

Therefore, in another setting, we evaluate images generated using our LLaVA-1.5-LLaMA3-8B recaptioned version of the raw COCO captions. Here, we observe consistent and significant improvements in both FID and CLIP scores, particularly when more than half of the recaptioned data are integrated into the training dataset. Notably, models trained on Recap-Datacomp-1B ($p=0$) surpass those trained on the vanilla Datacomp-1B ($p=1$) by a large margin, with improvements observed in FID (-8.4), CLIP score (+3.1), Recap-CLIP score (+8.4), and GPT-4V score (+1.1). These observations justify that Recap-Datacomp-1B better reveals the potential of text-to-image models in generating images with high visual quality and improved alignment with textual conditions.

\begin{figure}[t!]
    \centering
    \vspace{-1em}
    \includegraphics[width=0.9\linewidth]{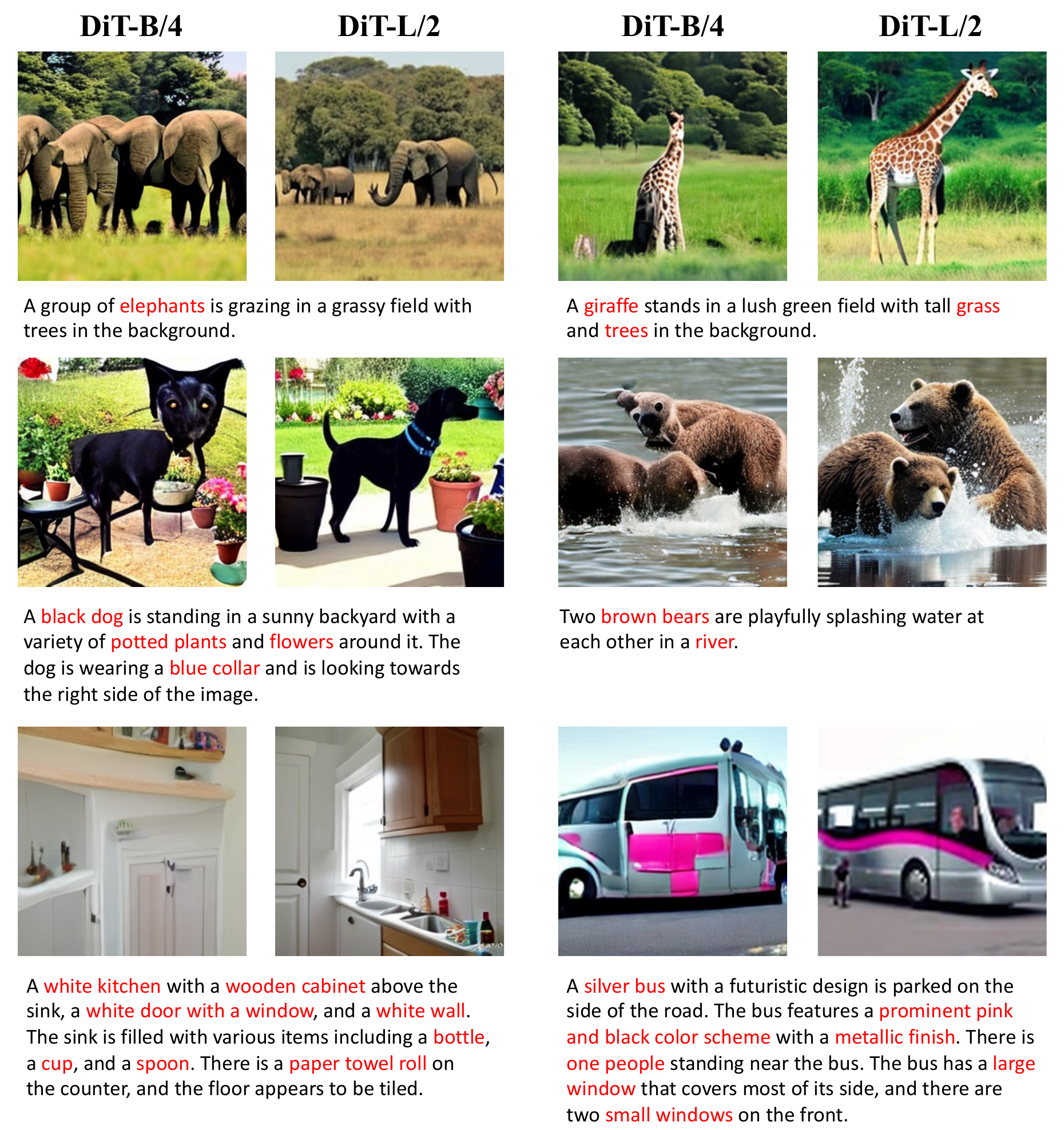}
    \caption{Visual comparison of generate results from  DiT-L/2 and DiT-B/4 at $p=0.0$, DiT-L/2 has better text comprehension and image generation than  DiT-B/4. We \textcolor{red}{mark} entities in the instruction.}
    \label{fig:Visual_comparison}
    \vspace{-1em}
\end{figure}

\textbf{Larger models.}
We further train a larger model, DiT-L/2, for 1 epoch with a mixed ratio of $p=0.0$, while keeping other training parameters unchanged. The model achieves an FID of 25.14 and a CLIP Score of 34.82. In Figure~\ref{fig:Visual_comparison}, we visually compare the generated results of DiT-L/2 and DiT-B/4 at $p=0.0$. It is evident that although the quantitative scores may not show substantial improvement, as we scale up the model, there is a noticeable enhancement in the alignment between the generated images and the corresponding text, \ie, this improved alignment results in higher-quality images that are able to capture and express more intricate details. These results confirm that DiT models trained on our recaption DataComp-1B exhibit robust scalability for text-to-image generative tasks.

\begin{table*}[t!]
\centering
\caption{Text-to-Image evaluation on COCO-30K results of DiT-BASE/4, trained with different mix ratios on Recap-DataComp-1B. Note for GPT-4V Score, we use a subset of 3K for the evaluation. }
\vspace{-.5em}
\resizebox{\linewidth}{!}{
\begin{tabular}{c|c|c|c|c|c|c}
\toprule
Training & \multicolumn{6}{c}{Evaluation} \\ \midrule
 \multirow{2}{*}{ mixed ratio $p$} & \multicolumn{2}{c|}{\textcolor{gray}{Raw}} & \multicolumn{4}{c}{Our COCO-Recap}  \\ \cmidrule{2-7}
  & \textcolor{gray}{FID$\downarrow$} & \textcolor{gray}{CLIP Score$\uparrow$}  & FID$\downarrow$ & CLIP Score$\uparrow$ & Recap-Clip Score$\uparrow$ & GPT-4V Score$\uparrow$ \\ 
\midrule
\cellcolor{mygray}0.00  & \cellcolor{mygray}\textcolor{gray}{37.6} & \cellcolor{mygray}\textcolor{gray}{29.2}  & \cellcolor{mygray}27.8\textcolor{red}{$_{-8.4}$} & \cellcolor{mygray}32.5\textcolor{red}{$_{+3.1\%}$} & \cellcolor{mygray}\textbf{28.3}\textcolor{red}{$_{+8.4\%}$} & \cellcolor{mygray}\textbf{2.53}\textcolor{red}{$_{+1.1}$} \\ 
0.05 & \textcolor{gray}{38.5} & \textcolor{gray}{29.1}  & 27.9 & 32.5 & 28.0 & 2.51 \\ 
0.10 & \textcolor{gray}{36.0} & \textcolor{gray}{29.7}  & \textbf{27.2} & 32.7 & 28.2 & 2.51 \\ 
0.15 & \textcolor{gray}{35.8} & \textcolor{gray}{29.9}  & 28.2 & \textbf{33.0} & 28.1 & 2.45 \\ 
0.20 & \textcolor{gray}{35.8} & \textcolor{gray}{29.8}  & 28.4 & 32.7 & 28.0 & \textbf{2.53} \\ 
0.50 & \textcolor{gray}{35.3} & \textcolor{gray}{29.3} & 30.2 & 31.9 & 26.7  & 2.13 \\ 
0.75 & \textcolor{gray}{31.3} & \textcolor{gray}{29.4} & 32.7 & 31.2 & 25.8 & 1.89 \\ 
1.00 & \textcolor{gray}{32.5} & \textcolor{gray}{28.9}  & 36.2 & 29.3 & 19.9 & 1.40 \\ 
\bottomrule
\end{tabular}
}
\label{tab:dit_mix_ratio}
\vspace{-1em}
\end{table*}

\section{Conclusion}
This paper introduces Recap-DataComp-1B, a large-scale image dataset paired with detailed textual descriptions, generated using the LLaMA-3-powered Llava model. Our comprehensive analysis reveals that, compared to the original, web-crawled textual data, these generated descriptions align more accurately with their corresponding images and are more detailed. Utilizing Recap-DataComp-1B for training resulted in consistent enhancements across various models, notably CLIP, particularly in image-to-text and text-to-image retrieval tasks, and in text-to-image Diffusion models, specifically in their ability to follow more closely to user-provided text instructions. By providing this high-quality, publicly available, large-scale image-text dataset, we hope to inspire ongoing research and development that will push the boundaries of developing vision-language foundation models, more particularly in the open-source community.

\subsection*{Acknowledge}
This work is partially supported by a gift from Adobe, TPU Research Cloud (TRC) program, Google Cloud Research Credits program, AWS Cloud Credit for Research program, Edinburgh International Data Facility (EIDF) and the Data-Driven Innovation Programme at the University of Edinburgh.

{\small
\bibliographystyle{plain}
\bibliography{main}
}

\end{document}